\def\year{2021}\relax
\documentclass[letterpaper]{article} 
\usepackage{aaai21}  
\usepackage{times}  
\usepackage{helvet} 
\usepackage{courier}  
\usepackage[hyphens]{url}  
\usepackage{graphicx} 
\usepackage{natbib}  
\usepackage{caption} 
\usepackage[switch]{lineno}
\usepackage{amssymb}
\usepackage{latexsym}
\usepackage{mathtools}
\usepackage{algorithm}
\usepackage[noend]{algpseudocode}
\usepackage{setspace}

\usepackage{multirow}
\usepackage{tabularx}
\usepackage{threeparttable}
\usepackage{booktabs}
\usepackage{stfloats}
\usepackage{subfigure}
\usepackage{colortbl}
\usepackage{xcolor}

\title{TAVAT: Token-Aware Virtual Adversarial Training for Language Understanding}
\author{
Linyang Li, Xipeng Qiu  \\
}
\affiliations{
Shanghai Key Laboratory of Intelligent Information Processing, Fudan University\\
School of Computer Science, Fudan University\\
\texttt{\{linyangli19, xpqiu\}@fudan.edu.cn}
}

\begin{document}

\maketitle

\begin{abstract}

Gradient-based adversarial training is widely used in improving the robustness of neural networks, while it cannot be easily adapted to natural language processing tasks since the embedding space is discrete.
In natural language processing fields, virtual adversarial training is introduced since texts are discrete and cannot be perturbed by gradients directly. 
Alternatively, virtual adversarial training, which generates perturbations on the embedding space, is introduced in NLP tasks.
Despite its success, existing virtual adversarial training methods generate perturbations roughly constrained by Frobenius normalization balls.
To craft fine-grained perturbations, we propose a Token-Aware Virtual Adversarial Training method.
We introduce a token-level accumulated perturbation vocabulary to initialize the perturbations better and use a token-level normalization ball to constrain these perturbations pertinently.
Experiments show that our method improves the performance of pre-trained models such as BERT and ALBERT in various tasks by a considerable margin.
The proposed method improves the score of the GLUE benchmark from 78.3 to 80.9 using BERT model and it also enhances the performance of sequence labeling and text classification tasks.

\end{abstract}

\section{Introduction}

Neural networks are proved vulnerable to crafted adversarial samples.
Recent works have shown that \textit{adversarial training} is helpful in constructing robust neural networks \cite{goodfellow2014explaining, Madry2018TowardsDL} against adversarial samples.
During adversarial training, gradients are collected from the clean samples to generate small perturbations. 
The gradients are collected and constrained by a normalization ball, then added to the original samples to create adversarial samples. 
Then these adversarial samples are used in the training process to improve the robustness against gradient-based adversarial attacks.

However, in the natural language processing fields, gradient-based adversarial attack and adversarial training methods cannot be easily adapted since the embedding space is discrete and gradients cannot be applied directly to form a perturbation.
Traditional methods used to generate adversarial samples for adversarial training usually incorporate substitution-based methods such as \citet{ebrahimi2017hotflip,Alzantot, jin2019textfooler, cheng2019robust, Li2020BERTATTACKAA}, which use synonyms or similar words/characters to replace the original ones in the sequence.
The process of these methods is inefficient since finding proper substitutions requires massive calculations.

In most cases, it is not necessary to craft substitution-based adversarial samples during the adversarial training process.
Instead, \citet{Miyato2017VirtualAT} propose a virtual adversarial training method that generates gradient-based perturbations on the embedding space as virtual adversarial samples,
which can help improve the generalization abilities of NLP models in various tasks \cite{miyato2016adversarial, Miyato2017VirtualAT, Zhu2020FreeLB:}. 

Despite the success of virtual adversarial training in NLP tasks, the perturbations generated from the virtual adversarial training method are rather rigid.
The most obvious problem is two-fold:

\textbf{(A) Initialization Problem}:

During the virtual adversarial training process, the perturbations are randomly initialized in every mini-batch.
Unlike the computer vision fields where pixels in an image do not possess information across instances, tokens in languages possess similar information in different sequences. 
Therefore, randomly initializing the perturbations on the same tokens in different sequences may neglect the fact that tokens could possess the same information and possible gradient direction across instances, which could be very helpful in generating fine-grained perturbations.
That is, random initialization of the perturbations could cause unnecessary noise during the virtual adversarial training.


\textbf{(B) Constraint Problem}:

On the other hand, traditional virtual adversarial training method constrains the perturbations with normalization balls which are usually Frobenius normalization.
Such a constraint generates an \textit{instance-level perturbation}:
the normalization ball acts on the embeddings of the entire sequence.
Therefore the constraint is not sensitive to different tokens in the sequence.
In natural languages, some of the tokens in the sequence play more important roles whiles others are relatively trivial in contributing valuable information to the task.
Therefore it is intuitive to construct token-level constraints to construct \textit{token-level perturbation} rather than \textit{sentence-level} constraints on the entire embedding space.

So in this paper, we propose a \textbf{Token-Aware} \textbf{V}irtual \textbf{A}dversarial \textbf{T}raining method that tackles the rigid perturbation construction used in previous virtual adversarial training methods.

To find a better initialization of the virtual perturbations of different tokens, we establish a \textit{global} perturbation vocabulary to accumulate the perturbations of the same tokens and use them as the perturbation initialization of the corresponding tokens.
In this way, we can avoid the noise caused by randomly initializing the perturbations of the same token in different sequences.
Therefore the perturbations are more pertinent in a global scale of the entire training samples of the given task.

To overcome the rigid Frobenius constraint used in the traditional virtual adversarial training process, we constrain the perturbations in the token-level.
We allow tokens with larger gradients to have a larger perturbation bound while tokens with trivial gradients are more tightly constrained. 
In this way the perturbations are fine-grained and more pertinent to different tokens.

Compared with substitution-based adversarial training methods, virtual adversarial training methods do not need to find the substitutes to replace the original sequences, so there are fewer constraints to keep the adversarial samples imperceptible.
Therefore, virtual adversarial training methods are much efficient.
Our \text{TA-VAT} method has no overhead compared with traditional virtual adversarial training method while generates fine-grained token-aware perturbations.

We construct extensive experiments to evaluate the effectiveness of these fine-grained token-aware virtual adversarial samples.
Results show that \text{TA-VAT} can boost the overall score of GLUE benchmark from 78.3 to 80.9 using the BERT-base model and from 89.9 to 90.9 using ALBERT xxlarge model. It is also helpful in sequence labeling tasks such as Conll 2003 and Ontonotes5.0 NER tasks. 

To summarize our contribution: 
we explore the detailed process of virtual adversarial training method used in NLP tasks and construct a fine-grained perturbation generation strategy named \textbf{TA-VAT}.
The proposed method has no overhead compared with previous virtual adversarial training methods while further improves the performances of NLP models in various kinds of tasks.

\section{Related Work}

\subsection{Adversarial Learning}
Adversarial attack \cite{goodfellow2014explaining} finds imperceptible perturbations to mislead neural networks.
In the computer vision field, adversarial attacks are extensively explored \cite{Carlini2017TowardsET} since it is easy to apply gradients over the continuous space in images.
Derived from gradient-based adversarial attacks, adversarial training \cite{goodfellow2014explaining} uses the generated adversarial samples to train the model against adversarial attacks. 
The PGD-algorithm \cite{Madry2018TowardsDL} is widely used in defense against adversarial attacks.
The PGD-algorithm process includes multiple projected gradient ascent steps to find the adversarial perturbations. Then these perturbations are used to update the model parameters.
Later developments such as \citet{freeat, Zhu2020FreeLB:} focus on finding better adversarial samples while maintaining a low calculation cost. 
That is to find adversarial perturbations and update the model parameters simultaneously in each gradient ascent step.

In the NLP fields, gradient-based methods would face a major challenge: texts are discrete, so gradients cannot be applied to the discrete tokens directly.
Instead of using gradients to craft perturbations, most common methods usually replace the original texts based on certain rules such as replacing words with semantically similar words \cite{Alzantot,jin2019textfooler, Li2020BERTATTACKAA}.
Besides replacing words, character-level and phrase-level adversarial samples are also introduced.
\citet{ebrahimi2017hotflip} proposes a perturbation strategy that can apply character insertion, deletion, and replacement.
\citet{jia2017adversarial} proposes a human-involved phrase generation method to mislead machine reading comprehension tasks.
These methods face a major challenge that finding the optimal solution in the massive space of possible combinations is hard.

Gradient-based methods such as \citet{ebrahimi2017hotflip,papernot2016crafting,cheng2019robust} used in generating adversarial samples in the texts domains usually find the substitutes that are similar to the gradient-based perturbations.
These replacing strategies cannot use normalization methods to constrain the perturbation imperceptible, so they use additional rules like synonym dictionaries or language models to measure whether the found perturbations are similar to the original samples.
These methods help improve the robustness of NLP models \cite{ebrahimi2017hotflip, Alzantot, jin2019textfooler}, or improve the performances of some NLP tasks such as machine translation \cite{cheng2019robust, cheng-etal-2020-advaug}.
However, finding a proper adversarial sample in the massive space of combinations through a trial and test process is usually costly when language models are involved in constraining the perturbation quality. 
So these methods are less efficient compared with the virtual adversarial training process.

\subsection{Virtual Adversarial Training Methods}
Virtual adversarial training methods \cite{miyato2016adversarial, Miyato2017VirtualAT} generate virtual adversarial samples in the embedding space as well as virtual labels, so these methods are helpful in tasks that do not require substitution-based adversarial samples.
The virtual adversarial training methods generate the perturbations based on gradients and constrain them with normalization balls on the embedding space, so the perturbations do not need to represent words/chars which are the so-called virtual adversarial samples, therefore these virtual adversarial training methods are different from gradient-based adversarial training methods such as \citet{cheng2019robust}.
Virtual adversarial training methods help improve the performances in semi-supervised text classifications \cite{Miyato2017VirtualAT} and help improve the generalization ability of pre-trained language models in downstream tasks \cite{Zhu2020FreeLB:}.

\section{Token-Aware Virtual Adversarial Training}

In this section, we first introduce the details of the gradient-based adversarial training/virtual adversarial training process, then we will illustrate the \textbf{Token-Aware VAT} method. 

\subsection{Gradient-Based Adversarial Training}

Normally, adversarial training aims to optimize parameter $\theta$ to minimize the maximum risk of misclassification when adding perturbations to the original inputs.
The perturbation $\delta$ is usually constrained by a norm ball $\epsilon$:
\begin{align}
     \mathop{min}\limits_{\theta} \mathbb{E}_{(\boldsymbol{X}, y)} \Bigg[\mathop{max}\limits_{||{\delta}||\leq \epsilon} L(f_{\theta}(\boldsymbol{X + \delta}), y) \Bigg]
     \label{eq:1}
\end{align}
where $y$ is the label of input $X$ and $L$ is the loss function of parameter $\theta$.
When it is a virtual adversarial training process, the input sequence $X$ is the embedding output of the input sequence.
Normally we use Frobenius norm to constrain $\delta$.

As pointed out by \citet{Madry2018TowardsDL}, in Equation \ref{eq:1}, the outer minimize function is non-convex, while the inner maximize function is non-concave.
A possible perturbation $\delta$ can be found through multiple steps of gradient ascent.
So in the PGD-algorithm, the perturbation $\delta$ is calculated by multiple steps of gradient ascent. 

At step $t$:
\begin{align}
    \boldsymbol{{\delta}}_{t+1} = {\prod}_{{||\boldsymbol{\delta}_t||}_F \leq \epsilon}
    \frac{(\boldsymbol{{\delta}}_{t} + \alpha g(\boldsymbol{{\delta}}_{t}))} {{||g(\boldsymbol{{\delta}}_{t})||}_F}
    \label{eq:2}
\end{align}
\begin{align}
    g(\boldsymbol{{\delta}}_{t}) = {\bigtriangledown}_{\boldsymbol{\delta}}L(f_{\theta}(\boldsymbol{X} + \boldsymbol{\delta}_t),y)
    \label{eq:3}
\end{align}
Here ${\prod}_{{||\boldsymbol{\delta}||}_F \leq \epsilon}$ represents the process that projects the perturbation onto the Frobenius normalization ball.

After multiple steps of gradient ascent, we acquire the perturbation $\delta$ and add it to the original inputs to train the given model.

\begin{algorithm*}[ht]
\setstretch{1.10}
\caption{Token-Aware Virtual Adversarial Training}\label{alg:sat}
\begin{algorithmic}[1]
\Require{Training Samples $S = \{(X=[w_0, \cdots, w_i, \cdots],y) \}$, perturbation bound $\epsilon$, initialize bound $\sigma$ adversarial steps $K$, adversarial step size $\alpha$, model parameter $\theta$}

\State $\boldsymbol{V} \in \mathbb{R}^{N \times D} \gets \frac{1}{\sqrt{D}} U(-\sigma, \sigma)$
 \textcolor[rgb]{0.00,0.50,0.00}{// Initialize perturbation vocabulary $\boldsymbol{V}$}
\For{epoch = $1, \cdots, $}

\For{batch $B \subset S$}
    
    \State $\boldsymbol{\delta}_{0}^{} \gets \frac{1}{\sqrt{D_{}}} U(-\sigma, \sigma)$
    , $\boldsymbol{\eta}_0^i \gets \boldsymbol{V}[w_i] $, $\boldsymbol{g}_0 \gets 0$ \textcolor[rgb]{0.00,0.50,0.00}{//Initialize perturbation and gradient of $\theta$}
    
    \For{t = $1, \cdots, K$}
        
        \State $\boldsymbol{g}_t \gets \boldsymbol{g}_{t-1} + \frac{1}{K} \mathbb{E}_{(X,y)\in B} [ \bigtriangledown_{\theta} L(f_{\theta}(X+\boldsymbol{\delta}_{t-1} + \boldsymbol{\eta}_{t-1}),y) ]$ \textcolor[rgb]{0.00,0.50,0.00}{//Accumulate gradients of $\theta$}

        \State \textcolor[rgb]{0.00,0.50,0.00}{Update token-level perturbation $\boldsymbol{\eta}$}:
        \State $\boldsymbol{g}_{\eta}^{i} \gets {\bigtriangledown}_{{\eta}^{i}} L(f_{\theta}((X+\boldsymbol{\delta}_{t-1} + \boldsymbol{\eta}_{t-1}),y)$
        \State $\boldsymbol{\eta}^i_{t} \gets 
        n^i * ( \boldsymbol{\eta}_{t-1}^i + \alpha \cdot \boldsymbol{g}_{\eta}^{i} / {||\boldsymbol{g}_{\eta}^{i}||}_{F} ) $

        \State $ \boldsymbol{\eta}_t^{} \gets {\prod}_{{|| {\boldsymbol{\eta}}_{}||}_F < \epsilon} (\boldsymbol{\eta}_t^{}) $

        \State \textcolor[rgb]{0.00,0.50,0.00}{Update instance-level perturbation $\boldsymbol{\delta}$}:

        \State $\boldsymbol{g}_{\delta}^{} \gets {\bigtriangledown}_{{\delta}^{}} L(f_{\theta}((X+\boldsymbol{\delta}_{t-1} + \boldsymbol{\eta}_{t-1}),y)$
        \State $\boldsymbol{\delta}^{}_{t} \gets {\prod}_{{||\boldsymbol{\delta}_{}||}_F < \epsilon}(\boldsymbol{\delta}_{t-1}^{} + \alpha \cdot \boldsymbol{g}_{\delta}^{} / {||\boldsymbol{g}_{\delta}^{}||}_{F}$)

    \EndFor
    \State {\textbf{end for}}
    
    \State $\boldsymbol{V}[w_i] \gets \boldsymbol{\eta}_K^i$ \textcolor[rgb]{0.00,0.50,0.00}{//Update perturbation vocabulary $\boldsymbol{V}$}
    
    \State $\theta \gets \theta - g_{K}$ \textcolor[rgb]{0.00,0.50,0.00}{//Update model parameter $\theta$}

\EndFor
\State {\textbf{end for}}
\EndFor
\State {\textbf{end for}}

\end{algorithmic}
\end{algorithm*}

\subsection{Token-Aware Virtual Adversarial Training}

As illustrated above, current virtual adversarial training methods craft virtual adversarial samples rigidly.
In NLP tasks, adversarial samples should be pertinent to different tokens since tokens in a sequence of natural languages may play different roles. 
Considering the two major omissions of current virtual adversarial training methods: randomly initialized perturbations and instance-level perturbation constraint, we propose the token-aware virtual adversarial training method to craft fine-grained token-aware virtual adversarial samples. 
The \textit{virtual adversarial training} concept we use represents that the adversarial samples are virtual, slightly different from \citet{Miyato2017VirtualAT}.

\textbf{Token-Level Perturbation $\boldsymbol{\eta}^i$}

The token-level perturbation is mainly two-fold: (1) initializing from a global perturbation vocabulary to avoid the noise caused by randomly initialization within the mini-batch; (2) constraining the perturbations at the token-level instead of a rigid normalization ball over the entire sequence.  

\begin{itemize}
    \item Global Perturbation Vocabulary:
    
    We create the global accumulated perturbation vocabulary $\boldsymbol{V} \in  \mathbb{R}^{N \times D} $, where $N$ is the vocabulary size and $D$ is the hidden dimension of the word embedding.
    In each minibatch, the token-level perturbation $\boldsymbol{\eta}^i_0$ of word $i$ is initialized by the corresponding perturbation from the global accumulated perturbation $\boldsymbol{V}$.
    After $K$ steps of gradient ascent, the corresponding words in the global accumulated perturbation $\boldsymbol{V}$ is updated by $\boldsymbol{\eta}^i_K$. 
    Therefore, in the next minibatch of the virtual adversarial training process, the perturbations can be initialized from the accumulated perturbation to avoid the unnecessary noise caused by the randomly initialized perturbations.
    
    \item Token-Level Constraints:
    
    After the initialization, we use gradients to update the perturbations and constrain them in a small normalization ball to keep the perturbations minimum. 
    However, the perturbations act on the embedding space which is token-level while they are normalized as an entire sequence in the previous virtual adversarial training process. 
    To bridge the gap between the token-level perturbation generalization and the instance-level perturbation constrain, we propose a token-level constraint.
    It is intuitive that different tokens play different roles in a sequence and some tokens may be vital to the task.     
    Therefore, we allow tokens with larger gradients to have larger perturbation bounds and restrict tokens with smaller gradients to have smaller bounds.
    Instead of using the Frobenius norm to normalize the entire perturbation, we normalize perturbations over separate tokens.
    Then we introduce a scaling index to allow larger perturbations on tokens with larger gradients.
    We calculate the scaling index $n^i$ by finding the maximum token-level perturbation in the sequence and set a scaling index according to that:
    \begin{align}
        n^i = \frac{{||{\boldsymbol{{\eta}}}^i_t||}_F}{\mathop{max}\limits_{j}({||{\boldsymbol{{\eta}}}^j_t||}_F)}
    \end{align}
    
    Finally, we still apply a Frobenius normalization over the scaled perturbation. 
    Therefore the token-level normalization constraint is formulated as:
    \begin{flalign}
        \boldsymbol{{\eta}}_{t+1}^i &=  n^i * \frac{(\boldsymbol{{\eta}}_{t}^i + \alpha g(\boldsymbol{{\eta}}_{t}^i))}
        {{||g(\boldsymbol{{\eta}}_{t}^i)||}_F)} \\
        \boldsymbol{{\eta}}_{t+1} &= {\prod}_{{||\boldsymbol{\eta}||}_F \leq \epsilon}(\boldsymbol{{\eta}}_{t})
    \end{flalign}
    In this way, the perturbations are flexible and pertinent to different tokens.

\end{itemize}

\textbf{Instance-Level Perturbation $\boldsymbol{\delta}$}

We adopt the instance-level perturbation in our token-aware virtual adversarial training algorithm as a complementary.  
We use $\delta$ to denote the instance-level perturbation which is the same used in illustrating the adversarial training process.
We calculate the instance-level perturbation $\delta$ using the adversarial training method illustrated in Equation \ref{eq:1}, \ref{eq:2} and \ref{eq:3}.

\textbf{Overall Process}

We illustrate the entire process in Algorithm \ref{alg:sat}.
We adopt the virtual adversarial training framework based on FreeLB \cite{Zhu2020FreeLB:}.
We first initialize the perturbation vocabulary. 
Instead of randomly initializing the perturbation, we initialize the token-level perturbation ${\eta}^i$ of the $i^{th}$ token using the corresponding token-level perturbation in the perturbation vocabulary $\boldsymbol{V}$.
During the training process, we calculate the gradients based on both instance-level perturbation $\delta$ and token-level perturbation ${\eta}^i$ as seen in line 7. 
We then calculate the gradients of token-level perturbation of the $i^{th}$ token ${\eta}^i$ and the gradients of the instance-level perturbation $\delta$ correspondingly.
We constrain the perturbations using the token-level normalization constraint with the scaling index. 

Finally, we update the perturbation vocabulary as well as the model parameters after the inner loop of gradient ascent steps.

\section{Experiments}

To evaluate the proposed \textbf{TA-VAT}, we construct extensive experiments over common NLP tasks: text classification, natural language inference, and named entity recognition.
We test on widely-used datasets: GLUE benchmark \cite{wang2019glue}, ConLL2003 NER dataset \cite{conll2003}, Ontonotes5.0 NER dataset \cite{weischedel2011ontonotes}, IMDB dataset and AG's NEWs dataset.

\subsection{Datasets}

\textbf{GLUE Dataset}

GLUE dataset is a collection of natural language understanding tasks, namely Multi-genre Natural Language Inference (MNLI \cite{mnli}); Quora Question Pairs (QQP \footnote{https://www.quora.com/q/quoradata/First-Quora-Dataset-Release-Question-Pairs}); Recognizing Textual Entailment (RTE \cite{rte}; Question Natural Language Inference (QNLI \cite{rajpurkar-etal-2016-squad}); Microsoft Research Paraphrase Corpus (MRPC \cite{dolan-brockett-2005-automatically}); Corpus of Linguistic Acceptability(CoLA \cite{warstadt2018neural}); Standard Sentiment Treebank (SST-2 \cite{socher-etal-2013-recursive}); Semantic Textual Similarity Benchmark (STS-B \cite{sts}.
All tasks except STS-B are formulated as a classification task.
STS-B is formulated as a regression task.

\textbf{NER Dataset}

Since our approach focuses on adversarial training concerning discrete tokens, we believe that such a method would improve the performance of sequence labeling tasks.
Therefore, we run NER task using CoNLL2003 dataset \cite{conll2003} and Ontonotes dataset \cite{weischedel2011ontonotes}.
The CoNLL 2003 dataset contains 12K training samples with 4 types of entities.
The Ontonotes dataset contains 60K training samples with 18 types of entities.

\textbf{Text Classification Dataset}

In the GLUE dataset, only SST-2 is a standard text classification task. We further run several popular classification datasets consists of news-genre classification and movie review classification which are longer sequences.
We use AG's NEWS dataset that predicts the news-type containing 112K training samples.
And we use the IMDB dataset \footnote{https://datasets.imdbws.com/}, a polarity sentiment classification task containing 45K training samples with an average length of 215 words.

\subsection{Implementations}

We implement our TA-VAT method with PyTorch based on Huggingface Transformers \footnote{https://github.com/huggingface/transformers}.
All models are trained using NVIDIA TitanXP GPUs.
We re-implement results of BERT, FreeAT, and FreeLB methods based on their open-released codes. 

We implement our approach based on hyper-parameters used in the standard fine-tuning process and the FreeLB adversarial training process.

Parameters such as the running epoch, learning rate, batch size and warmup step settings are the same as used in the standard fine-tuning process of BERT \footnote{https://github.com/google-research/bert} and ALBERT \footnote{https://github.com/google-research/albert/blob/master/}. 
As for hyper-parameters such as the adversarial training step $K$, the constrain bound of the perturbation $\epsilon$, the initialization bound $\sigma$ and the adversarial step size $\alpha$, we adopt parameters the same as used in FreeLB \footnote{https://github.com/zhuchen03/FreeLB} for a fair comparison. 
We adopt the same parameters when using the ALBERT-model, but we truncate some long sequences to save the GPU memories so the performances may be affected.

In the GLUE benchmark, we use the vanilla implementation of the original pre-trained language models.
We do not use the MNLI fine-tuned model to fine-tune the RTE task \cite{Zhu2020FreeLB:};
We formulate the QNLI task as a standard text classification task rather than a pairwise ranking task as a trick proposed by \citet{Liu2019RoBERTaAR}.
In the NER tasks, we use the cased BERT model \cite{bert}.

\begin{table*}[ht]\setlength{\tabcolsep}{4pt}
    \centering   
    \begin{tabular}{lcccccccccccc}
        \toprule

        \multirow{2}*{\bfseries Model}  & RTE & QNLI & MRPC & CoLA & SST & STS-B &MNLI-m/mm & QQP\\
        \cline{2-9} &Acc&Acc&Acc/f1&Mcc&Acc&P/S Corr&Acc&Acc/f1\\
        \midrule
        \multicolumn{3}{c}{\bfseries BERT-BASE}\\
        \midrule
        BERT \cite{bert} & -& 88.4 & -/86.7 & - & 92.7 & - & 84.4/- & - \\
        BERT-ReImp & 63.5 &91.1 & 84.1/89.0 & 54.7 & 92.9 & 89.2/88.8 &84.5/84.4 & 90.9/88.3\\
        FreeAT-ReImp  & 68.0&91.3 &85.0/89.2 & 57.5 & 93.2 & 89.5/89.0& 84.9 / 85.0 & 91.2/88.5\\
        FreeLB-ReImp  & 70.0&91.5 &86.0/90.0 & 58.9 & 93.4 & 89.7/89.2& 85.3 / 85.5 & 91.4/88.6\\
        TA-VAT(ours)  & \textbf{74.0} &\textbf{92.4} & \textbf{88.0/91.6} & \textbf{62.0} & \textbf{93.7} & \textbf{90.0}/\textbf{89.6}& \textbf{85.7} / \textbf{85.8} & \textbf{91.6}/\textbf{88.9 }\\
        \midrule
        \multicolumn{3}{c}{\bfseries ALBERT-xxlarge-v2}\\
        \midrule
        ALBERT-xxlarge-v2\cite{lan2019albert}  & 89.2 & 95.3 & -/90.9 & 71.4 & 96.9(96.5) & 93.0/- & 90.8/- & 92.2/-\\
        FreeLB\cite{Zhu2020FreeLB:} & 89.9 & 95.6$^*$ & -/92.4 & 73.1 & 97.0 & 93.2/- & 90.9/- & 92.5/- \\
        TA-VAT(ours) & \textbf{90.3} &  \textbf{95.7} & -/ \textbf{93.4} & \textbf{74.1} &  \text{96.8} &  \textbf{93.4}/- &  \textbf{91.1}/- &  \textbf{92.6} /-\\
        \bottomrule

    \end{tabular}
    \caption{Evaluation results on the development set of GLUE benchmark. 
    QNLI$^*$ in FreeLB is formed as pairwise ranking task.
}
    \label{tab:main}
\end{table*}

\begin{table*}[ht]\setlength{\tabcolsep}{6pt}
    \centering  
    \begin{tabular}{lcccccccccccc}
        \toprule

        \multirow{2}*{\bfseries Model} & RTE & QNLI & MRPC & CoLA & SST & STS-B&MNLI-m/mm & QQP \\

        \cline{2-9} &Acc&Acc&Acc/f1&Mcc&Acc&P/S Corr&Acc&Acc/f1\\

        \midrule
        BERT-BASE\cite{bert} & 66.4& 90.5 & 88.9/84.8 & 52.1 & 93.5 & 87.1/85.8& 84.6/83.4 &71.2/89.2  \\

        FreeLB\cite{Zhu2020FreeLB:} & 70.1& ${91.8}^*$ & 88.1/83.5 & 54.5 & 93.6 & 87.7/86.7& 85.7/84.6 &72.7/89.6\\
        TA-VAT(ours)  & \textbf{71.0}& \textbf{91.7} & \textbf{88.9/84.5} & \textbf{55.9} & \textbf{94.5} & 86.8/85.7 & 85.2/\textbf{84.7} &\textbf{72.8}/\text{89.5}\\

        \bottomrule

    \end{tabular}
    \caption{Evaluation results on the test set of GLUE benchmark.
    Results use the evaluation server on GLUE website.
    QNLI$^*$ in FreeLB is formed as pairwise ranking task.
}
    \label{tab:testsetglue}
\end{table*}

\subsection{Experiment Results}

As seen in Table \ref{tab:main}, \ref{tab:testsetglue}, \ref{tab:conll2003}, our Token-Aware Virtual Adversarial Training algorithm improves the fine-tuned models by a large margin.

Generally, TA-VAT lifts the evaluation dataset performance of the BERT-base model from 79.3 to 80.9.
The results are tested on the GLUE server \footnote{https://gluebenchmark.com/}.
The improvements indicate that the token-aware perturbations used in TA-VAT help construct fine-grained virtual adversarial samples to achieve greater generalization performances of pre-trained language models.

Compared with the FreeLB algorithm which does not incorporate token-aware adversarial training, our algorithm has a 0.4 points performance boost.
Further, TA-VAT can also boost performance when using the ALBERT model, which indicates that the method is effective in various pre-trained models. 
As mentioned, the xxlarge version of the ALBERT model costs large GPU memories so we truncate sequences to save the memories. The performances may be affected by a small margin.
In the SST dataset, we re-implement the results of the standard fine-tuning to 96.5 so the TA-VAT method does not surpass the reported performances of both the ALBERT model and the FreeLB model but outperforms the baseline of our implementation.

According to FreeLB \cite{Zhu2020FreeLB:}, a dropout strategy is used to boost the model performances by a considerable amount.
We do not explore how the dropout mechanism works in adversarial training, so we re-implement results of FreeLB without dropout for comparison.
The result of FreeLB-ALBERT in Table \ref{tab:main} is reported by \cite{Zhu2020FreeLB:}, which uses the dropout mechanism.
As seen in Table \ref{tab:main}, \ref{tab:testsetglue}, our method performs better even without the dropout mechanism.

In tasks like RTE or MRPC, the model performance is improved by a larger margin using TA-VAT.
We assume this is because the training sets in these tasks are limited to only a few thousands of samples, which indicates that our method can be more helpful in tasks that lack training data.
We will discuss the effectiveness of our method when dealing with insufficient data in the following section.

\begin{table}[ht]\setlength{\tabcolsep}{8pt}
    \centering  \small
    \begin{tabular}{lccccc}
        \toprule
        
        \multirow{1}*{\bfseries Model} &Pre & Recall & F1 \\
        \midrule
       \multicolumn{2}{c}{\bfseries CoNLL2003}\\
        BERT-ReImp & 94.9& 95.5 & 95.2 \\

        FreeLB-ReImp & 94.8 & 95.1  & 94.9\\
        TA-VAT(ours) & \textbf{95.0} & \textbf{95.7} & \textbf{95.4} \\
        \midrule
        \multicolumn{2}{c}{\bfseries Ontonotes}\\
        BERT-ReImp & 86.7& 88.9 & 87.8 \\
        FreeLB-ReImp & 86.7 &  89.2& 88.0\\
        TA-VAT(ours) & \textbf{87.0} & \textbf{89.5} & \textbf{88.2} \\
        \bottomrule

    \end{tabular}
    \caption{Evaluation results on the CoNLL2003 dataset and Ontonotes dataset.
}
    \label{tab:conll2003}
\end{table}

\begin{table}[ht]\setlength{\tabcolsep}{8pt}
    \centering  \small
    \begin{tabular}{lccccc}
        \toprule

        \multirow{1}*{\bfseries Model} &IMDB & AG's NEWS \\
        \midrule

        BERT-ReImp & 95.0& 90.0  \\
        FreeLB-ReImp & 95.4 & 90.5  \\
        TA-VAT(ours) & \textbf{95.7} & \textbf{90.9} \\

        \bottomrule

    \end{tabular}
    \caption{Evaluation results of the Text Classification Datasets.
}
    \label{tab:otherclassification}
\end{table}

In the sequence labeling tasks, the TA-VAT method can still boost the performance in both CoNLL 2003 NER dataset and the Ontonotes NER dataset. In CoNLL 2003 dataset, traditional adversarial training is even worse.
Therefore, we can assume that TA-VAT is effective in dealing with sequence-level tasks.

In the standard text classification tasks, TA-VAT is also effective.
In the IMDB dataset, the sequence is usually very long(over 200 words per sample).
As seen in \ref{tab:otherclassification}, the performance of TA-VAT is 0.3 points higher compared with the FreeLB method in the IMDB-dataset.

The results of various tasks indicate that the proposed TA-VAT method is effective in boosting the performance of fine-tuned pre-trained models.

\begin{table}[ht]\setlength{\tabcolsep}{8pt}
    \centering  \small
    \begin{tabular}{lccccc}
        \toprule
        \multicolumn{2}{c}{\bfseries Method} & RTE & MRPC & CoLA \\
        \bfseries Ptb-Vocab & \bfseries Tok-Norm  \\
        \midrule

        \checkmark & \checkmark & 74.0 & 88.0 & 62.0 \\
        - & \checkmark & 73.0 & 87.5 & 60.3 \\
        \checkmark &- & 72.5 & 87.0 & 59.5 \\
        - &- & 70.0 & 86.0 & 57.5 \\
        \bottomrule

    \end{tabular}
    \caption{Ablation Studies; Ptb-Vocab represents the perturbation vocabulary and Tok-norm represents the Token-level normalization constraint.
}
    \label{tab:ablation}
\end{table}

\subsection{Ablations}

We run ablation studies to explore the effectiveness of the key components in our adversarial training algorithm:

We setup ablation experiments to test the effectiveness of initializing the perturbations with the perturbation vocabulary.
Instead of initializing from the perturbation vocabulary, we initialize perturbation $\eta$ randomly for comparison.
Also, we run the experiment to compare whether using the token-level constraint is effective. 

As seen in Table \ref{tab:ablation}, without initializing the perturbations from the perturbation vocabulary, performances are considerably lower.
Also, using the token-level perturbation constraint is more effective than using the instance-level Frobenius norm to constrain the perturbations.

Therefore, we can summarize it is important to craft perturbations that concern the variance between tokens. 
Initializing the perturbations from a global accumulated perturbation vocabulary and using token-level constraints helps improve the quality of the generated virtual adversarial samples. 

\section{Analysis}

In this section, we construct experiments to further analyze the mechanism of the TA-VAT method.

\subsection{What does the perturbation vocabulary learn?}

Since we incorporate the global accumulated perturbation vocabulary during the training process, the perturbation vocabulary may learn some useful information after the entire training process.
So we construct an experiment: we save the perturbation vocabulary after the training process and add this vocabulary to the original word embedding layer of the pre-trained models. 
Then we run a normal fine-tuning process without the virtual adversarial training process on the same task using the updated pre-trained models.
We test on the development set using the BERT-base model.

Results in Table \ref{tab:ptbvocab} show that the learned perturbation vocabulary can help improve the performance of the simple fine-tuning method.
This indicates that the perturbation vocabulary helps improve the quality of the pre-trained word-embeddings during the fine-tuning stage.

Further, we update the word-embedding layer with the perturbation vocabulary learned in one task and run the normal fine-tuning process on another task to explore whether the perturbation vocabulary is transferable.

In Table \ref{tab:ptbvocab}, we can see that when the embedding is updated with another task, the performance of the normal fine-tuned model is also better than using the original embedding.
We assume that the perturbation vocabulary is transferable, which indicates that the learned perturbation vocabulary contains not only some task-specific information but also some universal information to improve the word-embeddings.

\begin{table}[ht]\setlength{\tabcolsep}{8pt}
    \centering  \small
    \begin{tabular}{lccccc}
        \toprule
        \multicolumn{1}{c}{\bfseries Method} & RTE & MRPC & CoLA \\
        \midrule
        
        Normal-Train & 63.5 & 84.1 & 54.7 \\
        TA-VAT & 74.0 & 88.0 & 62.0 \\
        Init-with-vocab(RTE) &  72.0 & 84.8 & 59.0 \\
        Init-with-vocab(MRPC) &  69.0 & 85.8 & 58.8 \\
        Init-with-vocab(CoLA) &  70.0 & 85.7 & 60.1 \\
        \bottomrule

    \end{tabular}
    \caption{Perturbation vocabulary: Init-with-vocab represents a normal fine-tuning process with embedding layer updated with the perturbation vocabulary learned in TA-VAT.
}
    \label{tab:ptbvocab}
\end{table}

\subsection{Does the special token play an important role?}

We accumulate the perturbations of certain words in the perturbation vocabulary, therefore, we can assume that some special tokens such as $[CLS]$ and $[SEP]$ in the BERT model may play vital roles in the training process.
We setup an experiment on the development set using BERT-base model to observe the performance that does not apply perturbations over these special tokens.

Results in Table \ref{tab:specialtoken} show that perturbations on the special tokens help improve the model performance.
Also, the perturbation vocabulary is not only useful in accumulating perturbations over the special tokens.

\begin{table}[ht]\setlength{\tabcolsep}{8pt}
    \centering  \small
    \begin{tabular}{lccccc}
        \toprule
        \multicolumn{2}{c}{\bfseries Method} & RTE & MRPC & CoLA \\
        \bfseries ST & \bfseries NT  \\
        \midrule

        \checkmark & \checkmark & 74.0 & 88.0 & 62.0 \\
         & \checkmark & 73.0 & 86.5 & 60.6 \\
        \checkmark &  & 73.5 & 87.0 & 61.0 \\
        \bottomrule

    \end{tabular}
    \caption{Perturbations of Special Tokens: ST is to include special tokens in the perturbation vocabulary; NT is to include normal tokens in the vocabulary.
}
    \label{tab:specialtoken}
\end{table}

\subsection{Does the TA-VAT improve robustness?}

Adversarial training was firstly introduced to improve the robustness against adversarial samples.
To explore whether our virtual adversarial training method helps improve model robustness,
we use the state-of-the-art adversarial attack algorithm Textfooler \cite{jin2019textfooler} to test the robustness of the trained model.
We use the IMDB dataset and test on the BERT-base model.
The implementation is the same as used in running the IMDB attack using Textfooler \footnote{https://github.com/jind11/TextFooler}.

As seen in Table \ref{tab:textfooler}, TA-VAT can improve the robustness against the Textfooler attacker.
The attacked accuracy is higher while the query number and the perturbation percentage are larger.
The query number is the number of access to the target model.
During the attack algorithm, the target model is accessed to return a score of the given texts for the attacker to modify its output adversaries.
This is a trial and test process so a larger query number indicates that the attacking process is harder.

\begin{table}[ht]\setlength{\tabcolsep}{4pt}
    \centering  \small
    \begin{tabular}{lccccc}
        \toprule
        \multicolumn{1}{c}{\bfseries Method} & Ori Acc & Atk Acc & Query Num & Ptb \\

        \midrule

        BERT & 90.9 & 13.6 & 1134 & 6.1\%\\

        TA-VAT & 98.9 & 15.8 & 2093 & 17.3\%\\
        \bottomrule

    \end{tabular}
    \caption{Adversarial attack results using Textfooler\cite{jin2019textfooler} as the attacker.
}
    \label{tab:textfooler}
\end{table}

\subsection{What task benefits more with TA-VAT?}

Experiments show that our TA-VAT algorithm is effective in various kinds of tasks, so it is also important to find out what kind of task benefits more with TA-VAT.
As already shown in Table \ref{tab:testsetglue}, our TA-VAT algorithm is more effective in dealing with RTE, CoLA tasks than MNLI and QQP tasks.
We intuitively believe that the corpus size of the task may be the cause of the performance difference.
Therefore, we construct an experiment that uses different proportions of the training set in the MNLI task to fine-tune the pre-trained models.

\begin{figure}[htbp]
\centering
\includegraphics[width=0.9\linewidth]{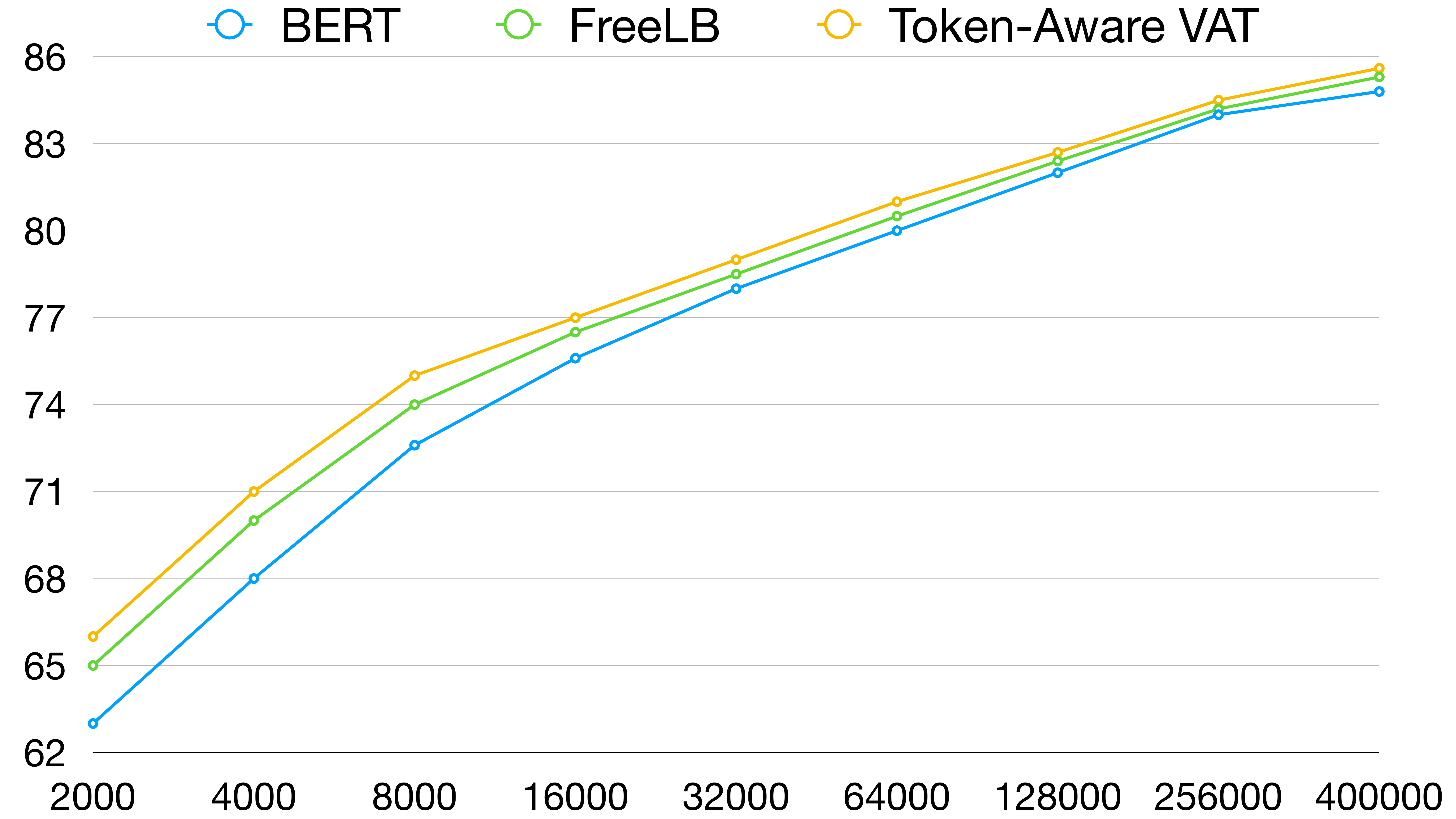}
\centering
\caption{Performance of MNLI Dataset trained with different training data size.}
\label{fig:corpus-size}
\end{figure}

As seen in Figure \ref{fig:corpus-size}, TA-VAT is more powerful when dealing with a relatively smaller training set.
When we train the MNLI task with only 2000 training pairs, TA-VAT can lift the performance by a larger margin than training with the full 400K dataset.

In the NLP field, obtaining a high-quality dataset is costly and most tasks have limited high-quality data.
We believe that our TA-VAT method can be widely used in these low-resource tasks.

\section{Conclusion}
In this paper, we focus on virtual adversarial training in the NLP field.
We propose a Token-Aware Virtual Adversarial Training method to allow virtual adversarial training methods to construct fine-grained virtual adversarial samples.
We establish experiments to show that our method helps improve the performance of various tasks using pre-trained language models.
In the future, we will further explore the potential of improving both generalization and robustness using token-aware virtual adversarial training methods.

\section{Acknowledgments}
We would like to thank the anonymous reviewers for their valuable comments. 
This work was supported by the National Key Research and Development Program of China (No. 2018YFB1005100).
We are thankful for the help of Hang Yan, Tianxiang Sun, Jiehang Zeng, Pengfei Liu and Ruotian Ma.

\bibliography{nlp.bib}

\begin{thebibliography}{28}
\providecommand{\natexlab}[1]{#1}
\providecommand{\url}[1]{\texttt{#1}}
\providecommand{\urlprefix}{URL }
\expandafter\ifx\csname urlstyle\endcsname\relax
  \providecommand{\doi}[1]{doi:\discretionary{}{}{}#1}\else
  \providecommand{\doi}{doi:\discretionary{}{}{}\begingroup
  \urlstyle{rm}\Url}\fi

\bibitem[{Agirre, M{\`a}rquez, and Wicentowski(2007)}]{sts}
Agirre, E.; M{\`a}rquez, L.; and Wicentowski, R., eds. 2007.
\newblock \emph{Proceedings of the Fourth International Workshop on Semantic
  Evaluations ({S}em{E}val-2007)}. Prague, Czech Republic: Association for
  Computational Linguistics.
\newblock \urlprefix\url{https://www.aclweb.org/anthology/S07-1000}.

\bibitem[{Alzantot et~al.(2018)Alzantot, Sharma, Elgohary, Ho, Srivastava, and
  Chang}]{Alzantot}
Alzantot, M.; Sharma, Y.; Elgohary, A.; Ho, B.; Srivastava, M.~B.; and Chang,
  K. 2018.
\newblock Generating Natural Language Adversarial Examples.
\newblock \emph{CoRR} abs/1804.07998.
\newblock \urlprefix\url{http://arxiv.org/abs/1804.07998}.

\bibitem[{Carlini and Wagner(2017)}]{Carlini2017TowardsET}
Carlini, N.; and Wagner, D.~A. 2017.
\newblock Towards Evaluating the Robustness of Neural Networks.
\newblock \emph{2017 IEEE Symposium on Security and Privacy (SP)} 39--57.

\bibitem[{Cheng, Jiang, and Macherey(2019)}]{cheng2019robust}
Cheng, Y.; Jiang, L.; and Macherey, W. 2019.
\newblock Robust neural machine translation with doubly adversarial inputs.
\newblock \emph{arXiv preprint arXiv:1906.02443} .

\bibitem[{Cheng et~al.(2020)Cheng, Jiang, Macherey, and
  Eisenstein}]{cheng-etal-2020-advaug}
Cheng, Y.; Jiang, L.; Macherey, W.; and Eisenstein, J. 2020.
\newblock {A}dv{A}ug: Robust Adversarial Augmentation for Neural Machine
  Translation.
\newblock In \emph{Proceedings of the 58th Annual Meeting of the Association
  for Computational Linguistics}, 5961--5970. Online: Association for
  Computational Linguistics.
\newblock \doi{10.18653/v1/2020.acl-main.529}.
\newblock \urlprefix\url{https://www.aclweb.org/anthology/2020.acl-main.529}.

\bibitem[{Dagan, Glickman, and Magnini(2005)}]{rte}
Dagan, I.; Glickman, O.; and Magnini, B. 2005.
\newblock The PASCAL Recognising Textual Entailment Challenge.
\newblock In \emph{Proceedings of the First International Conference on Machine
  Learning Challenges: Evaluating Predictive Uncertainty Visual Object
  Classification, and Recognizing Textual Entailment}, MLCW’05, 177–190.
  Berlin, Heidelberg: Springer-Verlag.
\newblock ISBN 3540334270.
\newblock \doi{10.1007/11736790_9}.
\newblock \urlprefix\url{https://doi.org/10.1007/11736790_9}.

\bibitem[{Devlin et~al.(2018)Devlin, Chang, Lee, and Toutanova}]{bert}
Devlin, J.; Chang, M.; Lee, K.; and Toutanova, K. 2018.
\newblock {BERT:} Pre-training of Deep Bidirectional Transformers for Language
  Understanding.
\newblock \emph{CoRR} abs/1810.04805.
\newblock \urlprefix\url{http://arxiv.org/abs/1810.04805}.

\bibitem[{Dolan and Brockett(2005)}]{dolan-brockett-2005-automatically}
Dolan, W.~B.; and Brockett, C. 2005.
\newblock Automatically Constructing a Corpus of Sentential Paraphrases.
\newblock In \emph{Proceedings of the Third International Workshop on
  Paraphrasing ({IWP}2005)}.
\newblock \urlprefix\url{https://www.aclweb.org/anthology/I05-5002}.

\bibitem[{Ebrahimi et~al.(2017)Ebrahimi, Rao, Lowd, and
  Dou}]{ebrahimi2017hotflip}
Ebrahimi, J.; Rao, A.; Lowd, D.; and Dou, D. 2017.
\newblock Hotflip: White-box adversarial examples for text classification.
\newblock \emph{arXiv preprint arXiv:1712.06751} .

\bibitem[{Goodfellow, Shlens, and Szegedy(2014)}]{goodfellow2014explaining}
Goodfellow, I.~J.; Shlens, J.; and Szegedy, C. 2014.
\newblock Explaining and harnessing adversarial examples.
\newblock \emph{arXiv preprint arXiv:1412.6572} .

\bibitem[{Jia and Liang(2017)}]{jia2017adversarial}
Jia, R.; and Liang, P. 2017.
\newblock Adversarial examples for evaluating reading comprehension systems.
\newblock \emph{arXiv preprint arXiv:1707.07328} .

\bibitem[{Jin et~al.(2019)Jin, Jin, Zhou, and Szolovits}]{jin2019textfooler}
Jin, D.; Jin, Z.; Zhou, J.~T.; and Szolovits, P. 2019.
\newblock Is {BERT} Really Robust? Natural Language Attack on Text
  Classification and Entailment.
\newblock \emph{CoRR} abs/1907.11932.
\newblock \urlprefix\url{http://arxiv.org/abs/1907.11932}.

\bibitem[{Lan et~al.(2019)Lan, Chen, Goodman, Gimpel, Sharma, and
  Soricut}]{lan2019albert}
Lan, Z.; Chen, M.; Goodman, S.; Gimpel, K.; Sharma, P.; and Soricut, R. 2019.
\newblock Albert: A lite bert for self-supervised learning of language
  representations.
\newblock \emph{arXiv preprint arXiv:1909.11942} .

\bibitem[{Li et~al.(2020)Li, Ma, Guo, Xue, and Qiu}]{Li2020BERTATTACKAA}
Li, L.; Ma, R.; Guo, Q.; Xue, X.; and Qiu, X. 2020.
\newblock BERT-ATTACK: Adversarial Attack Against BERT Using BERT.

\bibitem[{Liu et~al.(2019)Liu, Ott, Goyal, Du, Joshi, Chen, Levy, Lewis,
  Zettlemoyer, and Stoyanov}]{Liu2019RoBERTaAR}
Liu, Y.; Ott, M.; Goyal, N.; Du, J.; Joshi, M.; Chen, D.; Levy, O.; Lewis, M.;
  Zettlemoyer, L.; and Stoyanov, V. 2019.
\newblock RoBERTa: A Robustly Optimized BERT Pretraining Approach.
\newblock \emph{ArXiv} abs/1907.11692.

\bibitem[{Madry et~al.(2018)Madry, Makelov, Schmidt, Tsipras, and
  Vladu}]{Madry2018TowardsDL}
Madry, A.; Makelov, A.; Schmidt, L.; Tsipras, D.; and Vladu, A. 2018.
\newblock Towards Deep Learning Models Resistant to Adversarial Attacks.
\newblock \emph{ArXiv} abs/1706.06083.

\bibitem[{Miyato, Dai, and Goodfellow(2016)}]{miyato2016adversarial}
Miyato, T.; Dai, A.~M.; and Goodfellow, I. 2016.
\newblock Adversarial training methods for semi-supervised text classification.
\newblock \emph{arXiv preprint arXiv:1605.07725} .

\bibitem[{Miyato et~al.(2017)Miyato, ichi Maeda, Koyama, and
  Ishii}]{Miyato2017VirtualAT}
Miyato, T.; ichi Maeda, S.; Koyama, M.; and Ishii, S. 2017.
\newblock Virtual Adversarial Training: A Regularization Method for Supervised
  and Semi-Supervised Learning.
\newblock \emph{IEEE Transactions on Pattern Analysis and Machine Intelligence}
  41: 1979--1993.

\bibitem[{Papernot et~al.(2016)Papernot, McDaniel, Swami, and
  Harang}]{papernot2016crafting}
Papernot, N.; McDaniel, P.; Swami, A.; and Harang, R. 2016.
\newblock Crafting adversarial input sequences for recurrent neural networks.
\newblock In \emph{MILCOM 2016-2016 IEEE Military Communications Conference},
  49--54. IEEE.

\bibitem[{Rajpurkar et~al.(2016)Rajpurkar, Zhang, Lopyrev, and
  Liang}]{rajpurkar-etal-2016-squad}
Rajpurkar, P.; Zhang, J.; Lopyrev, K.; and Liang, P. 2016.
\newblock {SQ}u{AD}: 100,000+ Questions for Machine Comprehension of Text.
\newblock In \emph{Proceedings of the 2016 Conference on Empirical Methods in
  Natural Language Processing}, 2383--2392. Austin, Texas: Association for
  Computational Linguistics.
\newblock \doi{10.18653/v1/D16-1264}.
\newblock \urlprefix\url{https://www.aclweb.org/anthology/D16-1264}.

\bibitem[{Shafahi et~al.(2019)Shafahi, Najibi, Ghiasi, Xu, Dickerson, Studer,
  Davis, Taylor, and Goldstein}]{freeat}
Shafahi, A.; Najibi, M.; Ghiasi, A.; Xu, Z.; Dickerson, J.~P.; Studer, C.;
  Davis, L.~S.; Taylor, G.; and Goldstein, T. 2019.
\newblock Adversarial Training for Free!
\newblock \emph{CoRR} abs/1904.12843.
\newblock \urlprefix\url{http://arxiv.org/abs/1904.12843}.

\bibitem[{Socher et~al.(2013)Socher, Perelygin, Wu, Chuang, Manning, Ng, and
  Potts}]{socher-etal-2013-recursive}
Socher, R.; Perelygin, A.; Wu, J.; Chuang, J.; Manning, C.~D.; Ng, A.; and
  Potts, C. 2013.
\newblock Recursive Deep Models for Semantic Compositionality Over a Sentiment
  Treebank.
\newblock In \emph{Proceedings of the 2013 Conference on Empirical Methods in
  Natural Language Processing}, 1631--1642. Seattle, Washington, USA:
  Association for Computational Linguistics.
\newblock \urlprefix\url{https://www.aclweb.org/anthology/D13-1170}.

\bibitem[{Tjong Kim~Sang and De~Meulder(2003)}]{conll2003}
Tjong Kim~Sang, E.~F.; and De~Meulder, F. 2003.
\newblock Introduction to the CoNLL-2003 Shared Task: Language-Independent
  Named Entity Recognition.
\newblock In \emph{Proceedings of the Seventh Conference on Natural Language
  Learning at HLT-NAACL 2003 - Volume 4}, CONLL ’03, 142–147. USA:
  Association for Computational Linguistics.
\newblock \doi{10.3115/1119176.1119195}.
\newblock \urlprefix\url{https://doi.org/10.3115/1119176.1119195}.

\bibitem[{Wang et~al.(2019)Wang, Singh, Michael, Hill, Levy, and
  Bowman}]{wang2019glue}
Wang, A.; Singh, A.; Michael, J.; Hill, F.; Levy, O.; and Bowman, S.~R. 2019.
\newblock {GLUE}: A Multi-Task Benchmark and Analysis Platform for Natural
  Language Understanding.
\newblock In the Proceedings of ICLR.

\bibitem[{Warstadt, Singh, and Bowman(2018)}]{warstadt2018neural}
Warstadt, A.; Singh, A.; and Bowman, S.~R. 2018.
\newblock Neural Network Acceptability Judgments.
\newblock \emph{arXiv preprint arXiv:1805.12471} .

\bibitem[{Weischedel et~al.(2011)Weischedel, Pradhan, Ramshaw, Palmer, Xue,
  Marcus, Taylor, Greenberg, Hovy, Belvin et~al.}]{weischedel2011ontonotes}
Weischedel, R.; Pradhan, S.; Ramshaw, L.; Palmer, M.; Xue, N.; Marcus, M.;
  Taylor, A.; Greenberg, C.; Hovy, E.; Belvin, R.; et~al. 2011.
\newblock OntoNotes Release 4.0.
\newblock \emph{LDC2011T03, Philadelphia, Penn.: Linguistic Data Consortium} .

\bibitem[{Williams, Nangia, and Bowman(2018)}]{mnli}
Williams, A.; Nangia, N.; and Bowman, S. 2018.
\newblock A Broad-Coverage Challenge Corpus for Sentence Understanding through
  Inference.
\newblock In \emph{Proceedings of the Conference of the North American Chapter
  of the Association for Computational Linguistics: Human Language
  Technologies}, 1112--1122.

\bibitem[{Zhu et~al.(2020)Zhu, Cheng, Gan, Sun, Goldstein, and
  Liu}]{Zhu2020FreeLB:}
Zhu, C.; Cheng, Y.; Gan, Z.; Sun, S.; Goldstein, T.; and Liu, J. 2020.
\newblock FreeLB: Enhanced Adversarial Training for Natural Language
  Understanding.
\newblock In \emph{International Conference on Learning Representations}.
\newblock \urlprefix\url{https://openreview.net/forum?id=BygzbyHFvB}.

\end{thebibliography}

\end{document}